\documentclass[sigconf, authorversion]{acmart}
\renewcommand\footnotetextcopyrightpermission[1]{} 
\usepackage{booktabs} 

\setcopyright{none}

\usepackage{enumerate}
\usepackage{multirow}
\usepackage{bm}
\usepackage{array}

\makeatletter
\def\thickhline{%
  \noalign{\ifnum0=`}\fi\hrule \@height \thickarrayrulewidth \futurelet
   \reserved@a\@xthickhline}
\def\@xthickhline{\ifx\reserved@a\thickhline
               \vskip\doublerulesep
               \vskip-\thickarrayrulewidth
             \fi
      \ifnum0=`{\fi}}
\makeatother

\newlength{\thickarrayrulewidth}
\newcommand{\Short}{Shorter}
\newcommand{\short}{shorter}

\begin{document}
\title[A Framework for Authorial Clustering of \Short{} Texts]{A Framework for Authorial Clustering of \Short{} Texts in Latent Semantic Spaces}
\titlenote{We use the term ``\short'' in this paper to denote paragraph-long texts that are typically shorter than conventional documents.}

\author{Rafi Trad}
\orcid{0000-0002-5310-1281}
\affiliation{%
  \institution{Faculty of Computer Science, Otto von Guericke University Magdeburg}
  \streetaddress{Universitätsplatz 2}
  \city{Magdeburg}
  \state{Saxony-Anhalt}
  \postcode{39106}
}
\email{rafi.trad@ovgu.de}

\author{Myra Spiliopoulou}
\orcid{0000-0002-1828-5759}
\affiliation{%
  \institution{Faculty of Computer Science, Otto von Guericke University Magdeburg}
  \streetaddress{Universitätsplatz 2}
  \city{Magdeburg}
  \state{Saxony-Anhalt}
  \postcode{39106}
}
\email{myra@ovgu.de}

\renewcommand{\shortauthors}{R. Trad and M. Spiliopoulou}

\begin{abstract}

Authorial clustering involves the grouping of documents written by the same author or team of authors without any prior positive examples of an author's writing style or thematic preferences. For authorial clustering on \short{} texts (paragraph-length texts that are typically shorter than conventional documents), the document representation is particularly important: very high-dimensional feature spaces lead to data sparsity and suffer from serious consequences like the curse of dimensionality, while feature selection may lead to information loss. We propose a high-level framework which utilizes a compact data representation in a latent feature space derived with non-parametric topic modeling. Authorial clusters are identified thereafter in two scenarios: (a) fully unsupervised and (b) semi-supervised where a small number of \short{} texts are known to belong to the same author (must-link constraints) or not (cannot-link constraints).
	
We report on experiments with 120 collections in three languages and two genres and show that the topic-based latent feature space provides a promising level of performance while reducing the dimensionality by a factor of 1500 compared to state-of-the-arts. We also demonstrate that, while prior knowledge on the precise number of authors (i.e. authorial clusters) does not contribute much to additional quality, little knowledge on constraints in authorial clusters memberships leads to clear performance improvements in front of this difficult task. Thorough experimentation with standard metrics indicates that there still remains an ample room for improvement for authorial clustering, especially with \short{} texts.

\end{abstract}



\maketitle

\section{Introduction}
\label{sec:intro}
As users contribute news, opinions and arguments in social networks, it is important to ascribe each text to its author towards protecting an author's statements from misuse by others and also guarding the online community by advancing digital forensics. As pointed out by Stamatatos et al in \cite{Stamatatos2009Survey}, it is reasonable to assume that authors demonstrate distinct writing styles, which can be used for automated authorship analysis. Such an analysis includes two independent tasks, namely \textit{authorship verification} and \textit{authorial clustering}. 

Authorship verification takes as input a set of authors and a set of documents and assigns each document to an author, while authorial clustering assumes that information on authors of documents is unavailable or unreliable. Authorial clustering seeks to partition the set of documents into clusters such that each cluster corresponds to one author \cite{stamatatos2016clustering} \footnote{Stamatatos et al term the process of discerning the number of authors/clusters and then assigning the documents to clusters as ``complete authorial clustering'' \cite{stamatatos2016clustering}. Our approach is a complete authorial clustering approach, since we also derive the number of clusters. We use the terms ``complete authorial clustering'' and ``authorial clustering'' interchangeably and prefer the latter for brevity.}. 

Many authorial clustering approaches invest on advanced machine learning methods, like recurrent neural networks \cite{bagnall2016authorship}, word embeddings \cite{agarwal2019authorship} and sophisticated document representations \cite{bagnall2016authorship,gomez2018hierarchical,sari2016exploring} in a space with thousands of dimensions. High-dimensional feature spaces, however, tend to get sparser as the texts get shorter and suffer from consequences like the curse of dimensionality. In this work, we propose a new high-level framework for authorial clustering. Our attempt represents texts in a dense feature space derived with topic modeling and cluster them by authorship, utilizing unsupervised and a semi-supervised clustering algorithms.

We use non-parametric topic modeling so that it is unnecessary to know the number of topics spoken about beforehand. ``\short{} texts'', as used in this paper, indicate the kind of texts we may read in micro-blogs, online reviews and social media that are around one paragraph in length. These texts are \emph{shorter} than ordinary texts but are still long enough to remain practical for authorship analyses which are highly difficult even for humans \cite{rexha2018authorship}. In sum, our non-parametric topic modeling attempt represents texts as denser vectors in a low-dimensional feature space, the size of which is also automatically induced.

Authorial clustering encompasses the challenge of identifying the number of authors who have written the texts in the document collection under study \cite{stamatatos2016clustering}. For \short{} texts in the likes of social media and especially review platforms, this challenge is further exacerbated by the fact that some users contribute many texts, while others contribute very few, resulting in authorial clusters of substantially varying sizes. We address this issue parsimoniously: in a fully unsupervised way, estimating the number of authorial clusters, and in a semi-supervised way that uses a ``small'' number of instance-level must-link (ML) and cannot-link (CL) constraints. In other words, external pieces of information are capitalized on in the semi-supervised variant after they are manually provided by experts, where they say that documents \(x\) and \(y\) must belong (or mustn't belong) to the same (unknown) author. The first case resembles a ML constraint, and the latter a CL one.

Our contributions are as follows: (i) proposing the first high-level framework for authorial clustering in latent feature spaces whose size (i.e. number of latent factors) is also automatically inferred from the data themselves by non-parametric topic models; (ii) measuring the benefit we gain when we can annotate a small proportion of the data via semi-supervision, which capitalizes on minimal expert knowledge in the form of constraints and (iii) thorough multi-faceted evaluation of the framework against naive and state-of-the-art (SOTA) baselines on text collections of \short{} texts belonging to different languages and genres.

The evaluation results show that the unsupervised performance of the framework is comparable to the SOTA in a way, despite using 1500 times less dimensions. In general, the performance of all studied methods is modest, including SOTA, and marks an ample room for improvement to tackle authorial clustering effectively. Furthermore, few additional constraints boost the quality of authorial clusters in our efficient vector space to surpass SOTA.

The remainder of the paper is organized as follows: section \ref{sec:related} describes advances of authorial clustering, relevant works on authorship verification, and related literature on topic modeling. Our approach is presented in section \ref{sec:method}, followed by experimental design aspects (section \ref{sec:exp}) in which we also elaborate on the evaluation criteria and the baselines to which we compare our work. We close with a discussion and suggested improvements in section \ref{sec:conclusions}.

\section{Related Work}
\label{sec:related}
Authorial clustering was one of the 2016 and 2017 tasks of the PAN competition \footnote{https://pan.webis.de}. For this competition task, Bagnall et al proposed a character-level multi-headed recurrent neural network (RNN) as a language model and showed that this approach recognized authorship idiosyncrasies even with short texts discussing different topics \cite{bagnall2016authorship}.

Sari and Stevenson used k-means for authorial clustering, and investigated the potential of word embeddings in comparison to a tf-idf n-gram language model (at character level)  \cite{sari2016exploring}. They showed that the former is beneficial for multi-topic texts but it is also more computationally demanding without achieving substantially better performance  \cite{sari2016exploring}. Agarwal et al. went on to utilize word embeddings with tf-idf weights and employed hierarchical clustering algorithms to perform authorship clustering \cite{agarwal2019authorship}. Kocher and Savoy adopted a simple set of features of the most frequent terms (words and punctuation) to represent the authorship and writing styles, whereupon they constructed authorial clusters by establishing authorship links between authors and documents \cite{kocher2016unine}.

Garc{\'\i}a-Mondeja et al. used a Bag-of-Words language model with binary features, because the documents were too short to use frequencies, and then applied a $\beta$-compact algorithm that placed documents in the same cluster if they were maximally similar and more proximal than the threshold $\beta$ \cite{garcia2017discovering}.

G{\'o}mez-Adorno et al. used a hierarchical agglomerative clustering algorithm with average linkage and cosine similarity, and with the Calinski-Harabasz optimization criterion to determine the cut-off layer of the deprogram \cite{gomez2018hierarchical}. They considered two language models at term level, tf-idf and log entropy, and showed that the latter had best performance when used with the most frequent 20,000 terms in the vocabulary \cite{gomez2018hierarchical}. The aforementioned algorithms build clusters from vectors drawn on an elaborately crafted language model.

To the best of our knowledge, topic modeling has not been used for authorial clustering thus far, but it has been applied successfully for authorship verification \cite{potha2018intrinsic,potha2019improving,potha2019dynamic}. Potha and Stamatatos pointed out that the topic space to which they reduced the original high dimensional space was less sparse, less noisy and appropriate for language-independent learning \cite{potha2018intrinsic}.

In their follow-up work \cite{potha2019improving}, they combined different authorship verification methods with one of Latent Semantic Indexing (LSI) or Latent Dirichlet Allocation (LDA) and showed that topic modeling lifted up the performance \cite{potha2019improving}. The same authors used ensembles and showed that topic modeling leads to superior and robust models, even for scarce data \cite{potha2019dynamic}. Hence, the potential of low-dimensional document representation in the topic space \cite{potha2018intrinsic,potha2019improving} makes topic modeling worthy of investigation for authorial clustering as well. In addition, and similarly to authorial clustering, recent solutions to authorship verification invest on sophisticated text representations, as proposed by Ding et al. \cite{Ding2019Learning}, and on combinations of text representation and complex (dis)similarity functions, as proposed e.g. by Halvani et al. in \cite{halvani2017authorship}.

On the front of topic modeling, LDA, a mainstay therein, was introduced by Blei et al. as a 3-level hierarchical parametric Bayesian model for collections of discrete data, which models the generative probabilistic process that is believed to generate the data items, or documents \cite{LDA}. HDP (Hierarchical Dirichlet Processes) was introduced by Teh et al. to extend the well-known Latent Dirichlet Allocation (LDA) and capture the uncertainty pertaining to the number of topics via an additional Dirichlet process \cite{WhyeTeh2006_HDPs}.

It is worthy to note that conventional topic modeling, let it be parametric (e.g. LDA) or non-parametric (e.g. HDP), is generally used with long typical texts. Furthermore, Hong and Davison showed that LDA-based parametric topic models perform poorly on micro-blog texts \cite{Hong2010TM_Twitter}. LDA was extended by Rosen-Zvi et al. in their proposed Author-Topic Model (AT), which inferred topics and authors from large corpora simultaneously \cite{Rosen-Zvi:2010:LAM:1658377.1658381}; however, it was shown that the potential of AT didn't exceed that of standard LDA with short texts when LDA was trained on aggregated short texts \cite{Hong2010TM_Twitter}. An improvement in terms of the coherence and interpretability of inferred topics from very short texts, e.g. document titles, was achieved with the biterm topic model (BTM) by Yan et al. \cite{YanBTM2013} relying on co-occurrences of terms at corpus level rather than document level.

\section{Our Approach}
\label{sec:method}
\begin{figure*}[tb]
    \centering
    \includegraphics[height=1.5in]{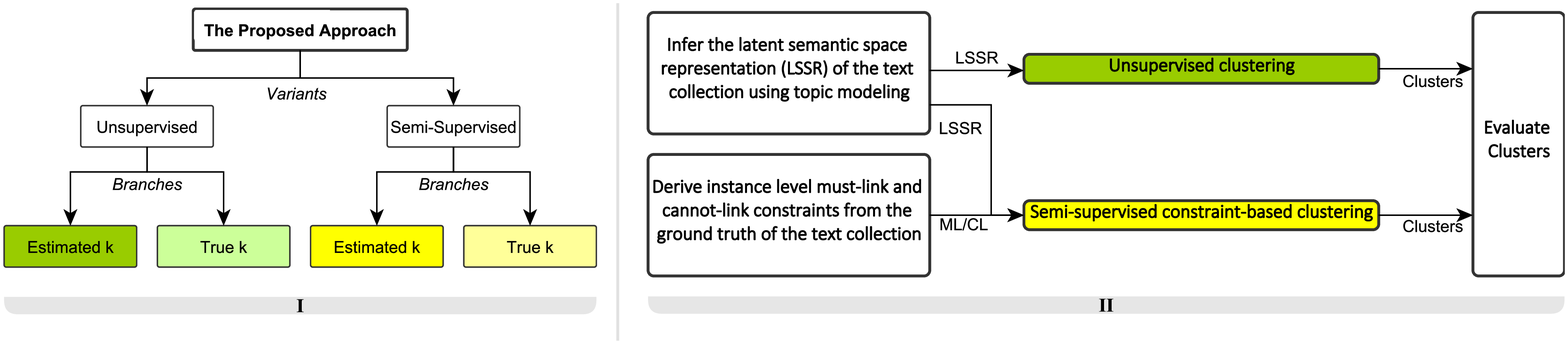}
    \caption{\textbf{An overview of our framework}. Part ``I'' on the left is a plain conceptual hierarchy of the framework's elements, and part ``II'' on the right summarizes the data flows.}
    \label{fig:methodology}
\end{figure*}{}

Generally, the proposed work encompasses: (1) learning a latent semantic representation (LSSR) of the texts as a low-dimensional, dense feature space after vectorizing on word unigrams. The number of latent topic-based dimensions is inferred via a non-parametric topic model. (2) Deriving authorial clusters in this feature space in two variants with respect to the learning model: fully unsupervised or semi-unsupervised with a restricted amount of background knowledge.

To test how good we can estimate the number of authors \(k\) in the corpus, we also supply the true \(k\) to each variant to test the differences in performance (figure \ref{fig:methodology}.I -- please note that \(k\) is different from number of topics \(t\)). Doing so isolates the effects of any suboptimal \(k\) estimations by usage of true \(k\) values instead, and thus monitors the severity of any consequent losses in performance.

As said, while LSSR is the only input to the unsupervised variant, the semi-supervised clustering variant takes advantage of a \emph{small} number of must-link and cannot-link (ML/CL) constraints to form the authorial clusters. This is stressed in (figure \ref{fig:methodology}.II) which highlights the data flow among our framework's tasks. Next we will describe how we construct the main input, i.e. LSSR.

\subsection{Building a dense feature space with non-parametric topic modeling}
The first step is to vectorize the documents. We use word unigrams which perform well in authorship verification settings \cite{potha2018intrinsic}, weighted by term frequencies. Furthermore, no punctuation is removed as it can contribute to the writing style of authors, and thus be valuable to discriminate authorial styles. On the other hand, too infrequent words which occur only once in a corpus are removed as they are most likely specific keywords that do not relate to the style of writing, and empirically we found this profitable while developing the approach with the dataset at hand. 

As mentioned, topic modeling is exploited afterwards as a dimensionality reduction technique to produce LSSR, the low-dimensional representation of texts with latent topics, which is also less noisy and language-independent. To this end, we use a non-parametric Bayesian topic model, namely Hierarchical Dirichlet Processes (HDP), implementing Gibbs sampling and inferring the number of topics $t$ automatically \cite{Yee2006HDP}. After this stage, each document \(d\) is represented as \(\vec{d_t}\) in terms of its weighted latent topics instead of observed words, and the weight of a topic in \(\vec{d_t}\) is the count of that topic's words in \(d\). \(\vec{d_t}\) characterizes \(d\) in the low dimensional latent semantic space which is manifested by the word-topic and document-topic matrices calculated by HDP.

\begin{table}[ht]
    \centering
    \caption{An example LSSR of a corpus embracing four documents, each of which contains \#w terms. LSSR is comprised of five dimensions in this example as per the number of ulterior inferred topics.}
    \label{tab:LSSR_Example}
    \begin{tabular}{|l|c||c|c|c|c|c|c|c|}
        \thickhline
        {\textbf{doc}} & \textbf{\#w} & {\textbf{t1}} & {\textbf{t2}} & {\textbf{t3}} & {\textbf{t4}} & {\textbf{t5}}\\
        \hline
        \textbf{d1} & 74 & 7 & 14 & 19 & 11 & 23 \\
        \hline
        \textbf{d2} & 50 & 8 & 11 & 9 & 12 & 10 \\
        \hline
        \textbf{d3} & 44 & 4 & 6 & 7 & 1 & 26 \\
        \hline
        \textbf{d4} & 60 & 11 & 15 & 7 & 15 & 12 \\
        \thickhline
    \end{tabular}{}
    \vspace{-3mm}
\end{table}{}

For example, table \ref{tab:LSSR_Example} illustrates an LSSR of a corpus of four documents, where five latent topics are identified automatically by HDP. We see that \(\vec{d1_t} = (7, 14, 19, 11, 23)\): \(d1\) contains 74 words, 7 of which belong to \(t1\), 14 to \(t2\) etc. In a similar fashion \(\vec{d2_t} = (8, 11, 9, 12, 10)\) and so on, and these vectors in the topic-based latent semantic space form inputs to subsequent clustering algorithms after they are l2-normalized on the sample level which renders each document a unit vector in the low-dimensional space. Such normalization neutralizes any induced biases by the length of texts and is encouraged by relevant literature \cite{banerjee2005spkmeansclustering} (We have also experimented with entropy-weighted logarithmic LSSR and found it comparable to the simpler l2-normalized LSSR, which we also adopt). We describe how \(k\), the number of authors, or clusters, is estimated next, followed by the very employed algorithms.

\subsection{Unsupervised authorial clustering variant}
\label{sec:k}
Determining the number of clusters \(k\) that governs a group of data points is a challenging task on its own, and it is an area of ongoing research. To this end, three estimation methods to estimate \(k\) are utilized: Optimization of intrinsic cluster evaluation metrics, the G-means algorithm \cite{hamerly2004learning} and the Gap statistic \cite{tibshirani2001estimating}.

\begin{enumerate}[I]
    \item \textbf{Optimizing Intrinsic Metrics}: This method finds an estimated \(k\) which optimizes an intrinsic clustering evaluation metric, like the average Silhouette coefficient, Calinski-Harabasz (CH) index or Davies–Bouldin index (DBI);
    \item \textbf{The G-means Algorithm}: built on the hypothesis that each resultant cluster in a proper clustering would conform to a Gaussian distribution \cite{hamerly2004learning}. We didn't test for the Gaussian assumption but we still made use of G-means to estimate \(k\);
    \item \textbf{The Gap Statistic}: A method introduced in \cite{tibshirani2001estimating} that can be seen as a formalization of the so-called "Elbow Method". The Gap statistic (Gap for short) can be computed using any clustering algorithm, and it compares the fluctuation in within-cluster dispersion with another expected dispersion under a pivot null distribution and across a series of values for \(k\).
\end{enumerate}{}

To complete our workflow for unsupervised authorial clustering, we couple the estimation process of $k$ with \emph{Spherical K-Means} algorithm whose performance is evaluated experimentally. Spherical K-Means can be regarded as the well-known K-Means but with cosine similarity \cite{banerjee2005spkmeansclustering}; which better suits the clustering of intrinsically directional data, like document vectors. As a side note, we also studied other clustering algorithms in this variant, but SPKMeans was selected due to its theoretical and practical fortes.

\subsection{Semi-supervised authorial clustering variant}
\label{ss_var}
The main difference between this variant and the unsupervised one is that we assume access to little expert knowledge in the form of constraints. We still need to estimate \(k\) but it would be estimated in a constraint-compliant manner by \emph{optimization of an intrinsic clustering evaluation method}, so a preparatory step to derive the instance-level ML/CL constraints is additionally needed (figure \ref{fig:methodology}.II). An ML constraint indicates that a pair of documents must belong to the same cluster, while a CL one states that a pair of documents cannot be related \cite{wagstaff2001constrained}. We did a small experiment and found CL constraints to be easier to define by us.

To identify ML/CL constraints, ground truths must be used to derive the constraints from. The number of constraints should be ``small'' enough to be practical, because in reality experts would need to annotate these texts. The detailed scheme whereby we elicit these constraints is described in section \ref{sssec:imp-details}.

We employ Constrained K-means, or COP K-Means \cite{wagstaff2001constrained} in this variant, which consumes the prior expert knowledge as must-link/cannot-link constraints  and thereby enhances the performance of the well-known K-Means \cite{wagstaff2001constrained}. Such an enhancement is especially helpful given the difficulty of the task at hand.

\section{Experimental Design}
\label{sec:exp}

In our experiments, we quantified the performance of the various configurations proposed within the framework, and compared them to naive and state-of-the-art baselines. Comparisons were made in terms of the quality of their clusterings against a true clustering, hence the usage of extrinsic evaluation metrics. As different methods to estimate \(k\) from LSSR were utilized, they were also measured against the true values of \(k\) in the analyzed corpora. The used corpora involved more than one language and genre, which shed light on how the performance of the methods fluctuated with respect to these aspects.

\subsection{The dataset}
\label{ssec:dataset}
PAN17-Clustering dataset \cite{tschuggnall2017overview} was used to measure the compared methods' performance, and we couldn't identify any other ready datasets which serve our purpose. PAN17-Clustering is publicly available\footnote{\url{https://pan.webis.de/data.html}} and comprises a number of clustering problems spanning three languages (English, Dutch and Greek) and two genres (articles and reviews). There are 60 authorial clustering problems reserved for development and tuning purposes and 120 problems for the final evaluation, allocated equally to the six language-genre combinations, i.e. 10 per each language-genre in the train segment and 20 in the test segment.

Each problem consists mostly of 20 single-authored texts in a specific language and genre, written by six authors on average \cite{tschuggnall2017overview}. Only the training dataset was used to tune the approach in question; no reference to the test dataset at any stage of development was made, which rendered our results comparable to the previous approaches that were applied to the same dataset.


The documents don't exceed 500 characters in length and are no less than 100. On average, English and Dutch articles exhibit \short{} texts than reviews whilst it is the contrary for Greek articles \cite{tschuggnall2017overview}. The majority of true authorial cluster sizes \(\in [1, 4]\), with singleton clusters pervading more than 25\% of the 1075 clusters in the dataset.

As this dataset is commonly used for authorship clustering, we want to re-state that English and Greek corpora were built by segmenting longer texts into paragraphs. For this reason, it is considerably easier to reassemble such texts again following some thematic hints and unique keywords/phrases found in the original document. Tschuggnal et al. acknowledge this fact \cite{tschuggnall2017overview} and we also observed this first-hand when attempting the task manually.

\subsection{Configurations and Baselines}
We compare our work to two baseline methods and to the winner of the PAN-2017 competition as the state-of-the-art \cite{gomez2018hierarchical,tschuggnall2017overview}. These are summarized in Table \ref{tab:allTheMethods}, and we describe the configurations next.

\begin{table*}[htb]
\centering
\caption{\protect{\label{tab:allTheMethods}}Overview of the methods used in our experiments. The prefix ``BL'' stands for ``baseline'', including SOTA.}
\begin{tabular}{|m{2.2cm}|m{1.6cm}|m{7.6cm}|m{5.0cm}|}
    \thickhline
    \textbf{Method's Acronym} & \textbf{Learning model} & \textbf{Description of the method} & \textbf{Estimation of $k$}
    \\ \hline
    BL\_r & - &\underline{B}ase\underline{L}ine from \cite{stamatatos2016clustering,tschuggnall2017overview}: it chooses a \underline{r}andom number as $k$ and builds $k$ clusters randomly & Random choice
    \\ \hline
    BL\_s & - &\underline{B}ase\underline{L}ine from \cite{stamatatos2016clustering,tschuggnall2017overview}: each document is a \underline{s}ingleton cluster & \(k = |Corpus|\)
    \\ \hline
    BL\_SOTA & Unsupervised & \underline{S}tate-\underline{O}f-\underline{T}he-\underline{A}rt as baseline: the winner of the PAN-2017 competition \cite{gomez2018hierarchical,tschuggnall2017overview}. & Optimal $k$ was chosen by the baseline itself through optimizing the Calinski-Harabasz (CH) index
    \\ \thickhline
    SPKMeans & Unsupervised & \underline{SP}herical \underline{KMeans} clustering elaborate method
    & Averaging out the estimations of G-Means and Gap
    \\ \hline
    COP-KMeans & Semi-Supervised & The semi-supervised constrained \underline{COP}-\underline{KMEANS} variant with 12\% of pairwise document links (ML/CL) revealed & Minimizing Davies–Bouldin index (DBI) penalized by \(k\) (i.e. \(DBI * k\))
    \\ \thickhline
\end{tabular}
\end{table*}

HDP topic model is in the core of our framework and generates the main inputs to SPKMeans and COP-KMeans variants. It was shown that HDP is on a par with the best selected LDA model through a model selection mechanism \cite{WhyeTeh2006_HDPs} and that the non-parametric methods outperform LDA with coupled model selection approaches \cite{Blei2004_HDP_nCRP}. For these reasons, we weren't encouraged to additionally implement our approach with LDA as a baseline, noting that it would need a model selection scheme or what's similar to determine \(t\).

We opted for Gibbs instead of Variational Inference (VI) for the reasons explained in \cite{Blei2017VI}: Gibbs is able to infer exact samples from the target density itself rather than from an approximate one, and the size of data is favorable for Gibbs, especially in our offline scenario. We conservatively set the number of iterations for the sampler to 10,000, twice the general recommended value in  \cite{raftery1991many}.

HDP has three hyperparameters, namely the topic Dirichlet prior $\eta$ and the concentration parameters $\gamma$ and $\alpha$. As pointed out by Wang and Blei in \cite{wang2012split}, these hyperparameters encode the priors of the inferred topics and thus control their density, which indirectly affects the number of topics and the dimensionality of LSSR. In addition to the default values $eta=0.5$ and $Gamma(1.0, 1.0)$ for $\gamma$ and $\alpha$, we investigated the effects of a dense, where $eta=0.8$, $\gamma$ and $\alpha$ = $Gamma(1.5, 1.0)$, and a sparse ($eta=0.3$, $\gamma$ and $\alpha$ = $Gamma(0.1, 1.0)$) settings. The sparse configuration yielded the highest per-word log likelihood in the training phase, so we used this configuration to infer LSSR.

One note before proceeding: we tried BTM instead of HDP in a proof of concept as it should produce more coherent and interpretable topics with extremely short texts \cite{YanBTM2013}. To this end, we used PAN-17 train data and sat \(t = 5\), and found the resultant authorial clusters worse than HDP's. As a possible explanation knowing that BTM relies on corpus-level words co-occurrences: the combination of terms that are not used together at document level seems counteractive for the task of finding documents adhering to the same authorial writing style. We use topic modeling as a means for intelligent reduction and densification of the sparse high-dimensional feature space in our context, and this task is orthogonal to the formation of interpretable, author-independent topics.

\subsubsection{Implementation details}
\label{sssec:imp-details}
In our high-level framework we used the original implementation of Hierarchical Dirichlet Process (HDP) with Gibbs sampling \cite{Yee2006HDP,wang2012split}\footnote{\url{https://github.com/blei-lab/hdp}}. Additionally, we utilized publicly available implementations of the core algorithms: G-Means 
\footnote{\url{https://github.com/flylo/g-means}}, Gap\footnote{\url{https://github.com/milesgranger/gap statistic}}, SPKMeans\footnote{\url{https://github.com/jasonlaska/spherecluster}} and COP-KMeans\footnote{\url{https://github.com/Behrouz-Babaki/COP-Kmeans}}. Using verified implementations makes our results more reliable and reproducible and saves a lot of development time which can be invested elsewhere.

As a note on the components which utilize randomness (HDP, COP-KMeans, SPKMeans, Gap and G-Means), we fixed the random seed in HDP and COP-KMeans but averaged out 5 full runs of SPKMeans which included running Gap and G-Means to estimate \(k\). We witnessed that doing so increased the reliability and reproducibility of SPKMeans results more than making the methods deterministic. Our code is written in Python 3 and is hereby made available\footnote{Code of the study: \url{https://github.com/rtrad89/authorship_clustering_code_repo}}.

\paragraph{Estimating \(k\)}
To determine \(k\) for Spherical K-Means, the estimates of G-means and Gap were averaged out and initial seeds were chosen with K-means++. On the other hand, a grid search in the range \([2, n-1]\), where \(n = |Corpus|\), was performed for Constrained K-means, and the selected \(k\) would minimize the quantity \(DBI * k\) (inspired by \cite{wagstaff2001constrained}) without making the algorithm fail.

\paragraph{Derivation of must-link and cannot-link constraints}
Must-link and cannot-link constraints serve as the prior expert knowledge on which Constrained K-means (COP-KMeans) can capitalize. Since we have the associated ground truth alongside every corpus in PAN17-Clustering dataset, we derived the aforesaid constraints therefrom, keeping the number of constraints ``small'' thus: Let there be \(n\) documents in a corpus, then we have \(l = \frac{n}{2}\times (n-1)\) pairwise links therein. We specified a small percentage (12\%) empirically by balancing improvement boost, feasibility and practicality, and after that we randomly sampled must-link and cannot-link constraints so that we cover 12\% of \(l\) without any influence on how many must-links or cannot-links we end up with.

\subsubsection{Evaluation Criteria}
\label{ssec:evalCriBase}
Extrinsic clustering evaluation measures are commonly used to assess the quality of clusterings when the ground truth is available. Amigo et al. systematically identified four useful formal constraints whereby the quality of a clustering evaluation metric can be judged \cite{Amigo2009BCubed} (\textit{Cluster homogeneity}, \textit{cluster completeness}, \textit{rag bag} and \textit{cluster size vs. quantity}), consequently advocating \(B^3F\) score as a measure that satisfies the aforementioned constraints. Additionally, we also extend the aforementioned set of constraints with another desirable property, the \textit{constant baseline property} \cite{vinh2010information}, which means the measure should be robust to random clusterings and be adjusted for chance. An obvious benefit is that unlike unadjusted metrics, adjusted ones do not tend to be biased towards a particular value of \(k\) - the number of clusters \cite{vinh2010information}.

As a result, \(\bm{B^3F}\) and \(\bm{ARI}\) scores were adopted in the evaluation framework of our experiments (using an open implementation\footnote{\url{https://github.com/hhromic/python-bcubed}} to calculate \(B^3F\) and Scikit-learn \cite{scikit-learn} for \(ARI\)). In addition, we list the mean rank \(\bm{\overline{r}}\) of the methods according to their \(B^3F\) and \(ARI\) scores. \(\overline{r}\) is a simple metric that averages out the six relative ranks (in the six language-genres of the test data), and it thus helps to also consider the consistency of exhibited performances when comparing the methods.

\subsection{Results}

We will contrast the performance of our proposed unsupervised and semi-supervised methods with state-of-the-art and naive baselines in terms of the evaluation measures. We will examine also the goodness of \(k\) estimation methods and how much loss in performance they incur, and lastly we briefly highlight some efficiency aspects.

\paragraph{The comparative study of performance levels}
``Estimated'' segments in tables \ref{tab:results} and \ref{tab:results_ARI} show the average $B^3F$ and $ARI$ of the methods in table \ref{tab:allTheMethods} respectively over the total of 120 test clustering problems. The expected \(\overline{B^3F}\) performance of BL\_SOTA as reported in \cite{tschuggnall2017overview} is \textbf{0.573}; although we use the same data and configuration as in \cite{gomez2018hierarchical} and don't change any preprocessing steps, we have observed lower $\overline{B^3F}$ scores on which we report.

\(\overline{B^3F}\) results indicate that our unsupervised variant is better than BL\_r and BL\_s, whilst the constrained-based variant COP-KMeans performs remarkably better. Additionally, the proposed methods are comparable to BL\_SOTA or better. COP-KMeans scores the best \(\overline{B^3F}\) at \emph{0.626} and ranks first, followed in rank by BL\_SOTA and closely by SPKMeans in that order. On the front of \(ARI\), the gap in performance between our unsupervised method and COP-KMeans increases; this is also the case when we compare it to BL\_SOTA. The constrained-based method of ours attains the highest \(ARI\) (\emph{0.335}) and also the best rank, followed by BL\_SOTA.

As explained in subsection \ref{ssec:dataset}, the data collection contains texts in three languages, organized according to two genres. In table \ref{tab:results}, we detail the \(B^3F\) performance of the methods on each genre in English, Dutch and Greek and show that several particularities have been suppressed by the average scores and the same goes for table \ref{tab:results_ARI} with \(ARI\). Articles whose writing style tends to be more structured elicit better \(B^3F\) and \(ARI\) results than reviews in general. Additionally, the simple BL\_r and BL\_s baselines have the lowermost $B^3F$ scores in all settings but Dutch reviews, where BL\_r shows competitive performance and BL\_s yields the best \(B^3F\). On the other hand, when we scrutinize table \ref{tab:results_ARI} we find that BL\_r and BL\_s have the lowest scores as \(ARI\) is adjusted for chance.

In terms of \(B^3F\), our COP-KMeans expectedly tops the scores in all sub-datasets but the Dutch documents, where it is still very competitive. In addition, our variants are particularly better than state-of-the-art's in Greek sub-dataset and especially articles. Since Dutch articles were built without segmentation in PAN-17 (cf. \ref{ssec:dataset}), they are the closest to reality, and SPKmeans surpasses SOTA in the reviews genre which also tends to be less structured usually, but not in Dutch articles.

Pertaining to English articles, SPKMeans performs better with articles than reviews. It seems to be less sensitive to differences in genre and language generally, and more consistent than the state-of-the-art; however, the semi-supervised COP-KMeans exhibits superior performance to all the other methods and baselines understandably.

\(ARI\) results also show a clear predominance of COP-KMeans whilst SPKmean is almost out of competition. COP-KMeans obtains the highest \(ARI\) in all sub-datasets but Greek and Dutch reviews, where the state-of-the-art BL\_SOTA achieves better scores but not by far. Moreover, BL\_SOTA scores higher \(ARI\) values than SPKMeans.

\begin{figure*}[!htp]
    \vspace{-4mm}
    \centering
    \includegraphics*[scale=0.6, trim= 40 50 5 50]{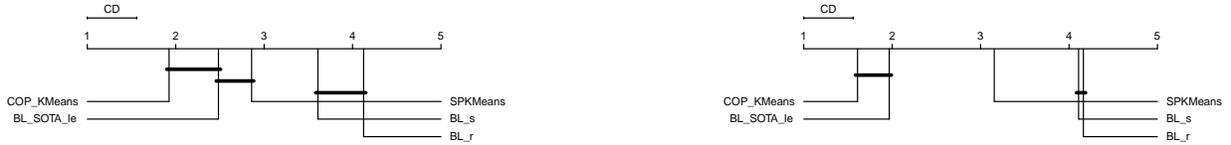}
    \caption{The Critical Difference diagrams of the competing methods according to their \(B^3F\) (left) and \(ARI\) (right) scores on the 120 test problem sets. Methods are ordered from best (leftmost) to worst (rightmost) and those that are not significantly different (at \(\alpha_{corrected} = 0.05\)) are connected with horizontal lines.}
    \label{fig:cd.plots}
\end{figure*}

\paragraph{Statistical significance tests} We should test \(B^3F\) and \(ARI\) scores which correspond to ``Estimated'' segments in tables \ref{tab:results} and \ref{tab:results_ARI} to substantiate that the observed differences in performance are actually significant. The data do not satisfy normality and homoscedasticity, so a non-parametric post-hoc test should be adopted. We use the conservative Friedman-Nemenyi non-parametric post-hoc tests and set \(\alpha_{corrected} = \frac{0.1}{2}\); Bonferroni correction is applied due to the small sample size \(N\) of the test \cite{Labovitz1968siglevel}. Friedman's Aligned Ranks test confirms that there are significant performance differences at \(\alpha_{corrected}\), and post-hoc Nemenyi tests show that our methods and SOTA demonstrate a significantly better performance than the naive baselines BL\_r and BL\_s in all evaluation metrics. COP-KMeans is not significantly different from the state-of-the-art BL\_SOTA in both metrics, while SPKMeans was significantly different from SOTA in \(ARI\) but not in \(B^3F\). Figure \ref{fig:cd.plots} details the results of the post-hoc tests using critical difference (CD) plots in terms of \(B^3F\) (left) and \(ARI\) (right).

\paragraph{Quality of \(k\) estimations}
\(k\) had to be estimated in SPKMeans and COP-KMeans. To investigate the differences between the estimated and true $k$ values for the different problems, we compute the Root Mean Squared Error (RMSE) over all problem sets. The results are shown in table \ref{tab:rmse} (SPKMeans, Gap and G-means RMSEs correspond to the mean of estimations obtained from five SPKMeans runs, as said earlier). The values indicate that the estimated numbers of clusters in the test dataset differ from the true number (which lies in [2, 10] and average at 6) by at least 2 (clusters). Notably, COP-KMeans constraint-compliant estimations are the closest to the associated real \(k\) values, followed by SPKMeans.

Segments ``True'' in tables \ref{tab:results} and \ref{tab:results_ARI} show the performance of our relevant methods when the number of authorial clusters $k$ is known, and they consequently monitor the loss in performance due to estimating \(k\) and deviating from its true value. In terms of \(B^3F\), SPKMeans and COP-KMeans could capitalize on knowing the true \(k\) values slightly. Nonetheless, supplying the true \(k\) values to the relevant methods hardly changes \(ARI\)'s outcomes.

\paragraph{Efficiency aspects} The slowest procedure in our approach is generating LSSR. There are around 20 documents of no more than 500 characters each in a problem set, and on average, it takes \(26.84\pm1.8(s_{\overline{x}})\) seconds to represent them as LSSR. Our LSSR utilizes \emph{no more than 13} latent dimensions to represent documents across all corpora, while the state-of-the-art uses 20,000 features \cite{gomez2018hierarchical}. As a result, executing clustering algorithms in LSSR takes a relatively negligible amount of time.

\begin{table*}[!htp]
    \centering
    \caption{\textbf{The overall mean rank \(\overline{r}\) (lower is better) and averaged \(B^3 F\) scores (higher is better) of the different approaches over the 120 test problem sets}. The \(B^3 F\) results are also detailed on the language-genre level where en: English, gr: Greek, nl: Dutch; ar: Articles and re: Reviews. k acquisition (abbreviated as acq.) column segments the table according to how \(k\) is provided to algorithms.}
    \begin{tabular}{|c|l|c|c||c|c|c|c|c|c|}
        \hline
        & & & \multicolumn{7}{c|}{\(\bm{B^3F}\)} \\
        \cline{4-10}
        \textbf{k acq.} & \textbf{algorithm} & \textbf{\(\bm{\overline{rank}}\)} & {overall mean} & {en-ar} & {en-re} & {gr-ar} & {gr-re} & {nl-ar} & {nl-re} \\
        \hline \hline
        \multirow{5}{*}{Estimated} & BL\_r & 4.7 & 0.435 & 0.466 & 0.444 & 0.411 & 0.418 & 0.428 & 0.441 \\
        \cline{2-10}
         & BL\_s & 3.8 & 0.458 & 0.436 & 0.475 & 0.403 & 0.455 & 0.438 & \textbf{0.543} \\
        \cline{2-10}
        & BL\_SOTA\_le & 2.3 & 0.562 & 0.602 & 0.570 & 0.532 & 0.535 & \textbf{0.670} & 0.461 \\
        \cline{2-10}
         & SPKMeans & 2.8 & 0.542 & 0.556 & 0.552 & 0.569 & 0.514 & 0.553 & 0.507 \\
         \cline{2-10}
         & COP-KMeans & \textbf{\textit{1.3}} & \textbf{\textit{0.626}} & \textbf{0.682} & \textbf{0.635} & \textbf{0.676} & \textbf{0.590} & 0.643 & 0.530 \\
        \hline \hline
        \multirow{2}{*}{True} & SPKMeans & - & 0.562 & 0.574 & 0.569 & 0.580 & 0.530 & 0.579 & 0.540 \\ 
        \cline{2-10}         
         & COP-KMeans & - & 0.649 & 0.700 & 0.654 & 0.709 & 0.603 & 0.642 & 0.585 \\
        \hline
    \end{tabular}
    \label{tab:results}
\end{table*}{}

\begin{table*}[!htp]
    \centering
    \caption{The overall mean rank \(\overline{r}\) and averaged \(ARI\) scores of the different approaches over the 120 test problem sets. We use the same notation described in table \ref{tab:results}.}
    \makebox[\textwidth][c]{
    \begin{tabular}{|c|l|c|c||c|c|c|c|c|c|}
        \hline
        & & & \multicolumn{7}{c|}{\(\bm{ARI}\)} \\
        \cline{4-10}
        \textbf{k acq.} & \textbf{algorithm} & \textbf{\(\bm{\overline{rank}}\)} & overall mean & {en-ar} & {en-re} & {gr-ar} & {gr-re} & {nl-ar} & {nl-re} \\
        \hline \hline
        \multirow{5}{*}{Estimated} & BL\_r & 4.5 & -0.001 & 0.006 & 0.016 & -0.028 & -0.003 & -0.014 & 0.017
        \\ \cline{2-10}
        & BL\_s &  4.5 & 0.000 & 0.000 & 0.000 & 0.000 & 0.000 & 0.000 & 0.000
        \\ \cline{2-10}
        & BL\_SOTA\_le & 1.7 & 0.265 & 0.282 & 0.288 & 0.215 & \textbf{0.293} & 0.315 & \textbf{0.197}
        \\ \cline{2-10}
        & SPKMeans & 3 & 0.100 & 0.115 & 0.091 & 0.135 & 0.082 & 0.102 & 0.076 \\
        \cline{2-10}
         & COP-KMeans & \textbf{\textit{1.3}} & \textbf{\textit{0.335}} & \textbf{0.439} & \textbf{0.361} & \textbf{0.420} & 0.259 & \textbf{0.389} & 0.141 \\
        \hline \hline
        \multirow{2}{*}{True} & SPKMeans & - & 0.108 & 0.132 & 0.096 & 0.140 & 0.085 & 0.119 & 0.077 \\ 
        \cline{2-10}
         & COP-KMeans & - & 0.338 & {0.444}  & {0.365}  & {0.452}  & 0.252  & {0.330}  & 0.182 \\
        \hline
    \end{tabular}
    } 
    \label{tab:results_ARI}
\end{table*}


\begin{table}[b]
    \centering
    \caption{\textbf{RMSE values of the \(k\) estimations} (lower is better). True values of k fall in \([2 , 10]\) with \(\overline{x} = 5.99\pm1.9(s)\) in test data.}
    \begin{tabular}{|l||c|l||c|}
    \hline
        BL\_SOTA: & \multicolumn{3}{l|}{3.379}\\
        
        \thickhline
        
        COP-KMeans: & \textbf{\textit{2.170}} &
        SPKMeans: & 2.664 \\ \hline
        
        \textcolor{gray}{Gap:} &\textcolor{gray}{4.541} &
        \textcolor{gray}{G-Means:} & \textcolor{gray}{3.771} \\ \hline
        
    \end{tabular}
    \label{tab:rmse}
\end{table}{}

{}

\section{Discussion and Conclusion}
\label{sec:conclusions}
This work is the first to systematically study non-parametric topic modeling towards authorial clustering of paragraph-long shorter texts via a high-level framework. By measuring the cosine similarity among documents in LSSR, the introduced approach can also be easily extended to support \textit{authorship-link ranking} \cite{stamatatos2016clustering} IR applications, in which documents are to be retrieved and ranked by authorship.

Conventionally, related works are measured in terms of \(B^3F\), but we also used a standard adjusted-for-chance metric (\(ARI\)) to judge the performance with regard to many aspects: \(B^3F\) provides an assessment attentive to suboptimal clusterings and \(ARI\) augments the evaluation framework with its adjusted-for-chance nature. We found that \(B^3F\) is more sensitive and informative than \(ARI\) and we tend to explain that by the fact that \(B^3F\) respects differences even among suboptimal results, which is the level we are generally at in this demanding application area.

There were distinctive performance differences among the compared methods, and the semi-supervised COP-KMeans topped the charts on all evaluation fronts, making us say it was thus the best performing method. Moreover, its constraint-based \(k\) estimator was the best in terms of RMSE and it remained close enough to true \(k\) values which didn't cripple the performance of the algorithm or make it suffer from extreme under-/over-estimations. It was significantly better than the unsupervised variant SPKMeans, and while it is indeed not directly comparable to SOTA, \emph{this suggests that when operating in low-dimensional topic-based vector spaces, capitalizing on little prior expert knowledge wherever applicable is considerably beneficial for authorial clustering of shorter texts}.

When focusing on SPKMeans vs. BL\_SOTA, it is clear that the latter reached higher \(\overline{B^3F}\) and \(\overline{ARI}\). We still want to note that: (1) Both methods significantly surpassed the naive baselines and produced comparable \(B^3F\) scores, and (2) SPKMeans with LSSR alleviates the need to engineer and tune down the dimensionality of the feature space because it operates in a dynamically-inferred lower dimensional space (the maximum number of topics whereby test documents were represented was \emph{13}; that's 1500x less than BL\_SOTA's 20000 dimensions). These two points hint that the presented methodology is worthy of further consideration.

Furthermore, we demonstrated that \(ARI\) scores of all methods were low, and no tested method, including the state-of-the-art, achieved an acceptable \(ARI\) in absolute terms. This fact is also highlighted implicitly by \(B^3F\) if we take BL\_r and BL\_s into account, which indicates the difficulty of authorial clustering with \short{} texts, and marks the need for more substantial work. For this reason and to identify a feasible next step, we attempted the task by hand and found that we could have a say on the (dis)similarity of some of the texts, thereby we suggest to focus future works on \emph{similarity/metric learning} first, to devise a similarity function that discerns how similar two texts are in stylometric terms, and then authorial clustering can proceed using that specialized function.

Moreover, cluster ensembles proved effective for authorship verification \cite{potha2019dynamic}, a very close task to authorial clustering, thus we deem them a viable option. Besides, crafting additional datasets where instances are completely unrelated is a sound next step to increase the reliability and reproducibility of any experimentation.

\bibliographystyle{ACM-Reference-Format}
\bibliography{pub} 


\begin{thebibliography}{33}


\ifx \showCODEN    \undefined \def \showCODEN     #1{\unskip}     \fi
\ifx \showDOI      \undefined \def \showDOI       #1{#1}\fi
\ifx \showISBNx    \undefined \def \showISBNx     #1{\unskip}     \fi
\ifx \showISBNxiii \undefined \def \showISBNxiii  #1{\unskip}     \fi
\ifx \showISSN     \undefined \def \showISSN      #1{\unskip}     \fi
\ifx \showLCCN     \undefined \def \showLCCN      #1{\unskip}     \fi
\ifx \shownote     \undefined \def \shownote      #1{#1}          \fi
\ifx \showarticletitle \undefined \def \showarticletitle #1{#1}   \fi
\ifx \showURL      \undefined \def \showURL       {\relax}        \fi
\providecommand\bibfield[2]{#2}
\providecommand\bibinfo[2]{#2}
\providecommand\natexlab[1]{#1}
\providecommand\showeprint[2][]{arXiv:#2}

\bibitem[\protect\citeauthoryear{Agarwal, Thakral, Bhatt, and Mittal}{Agarwal
  et~al\mbox{.}}{2019}]%
        {agarwal2019authorship}
\bibfield{author}{\bibinfo{person}{Lucky Agarwal}, \bibinfo{person}{Kartik
  Thakral}, \bibinfo{person}{Gaurav Bhatt}, {and} \bibinfo{person}{Ankush
  Mittal}.} \bibinfo{year}{2019}\natexlab{}.
\newblock \showarticletitle{Authorship Clustering using TF-IDF weighted
  Word-Embeddings}. In \bibinfo{booktitle}{\emph{Proceedings of the 11th Forum
  for Information Retrieval Evaluation}}. \bibinfo{pages}{24--29}.
\newblock


\bibitem[\protect\citeauthoryear{Amig{\'o}, Gonzalo, Artiles, and
  Verdejo}{Amig{\'o} et~al\mbox{.}}{2009}]%
        {Amigo2009BCubed}
\bibfield{author}{\bibinfo{person}{Enrique Amig{\'o}}, \bibinfo{person}{Julio
  Gonzalo}, \bibinfo{person}{Javier Artiles}, {and} \bibinfo{person}{Felisa
  Verdejo}.} \bibinfo{year}{2009}\natexlab{}.
\newblock \showarticletitle{A comparison of extrinsic clustering evaluation
  metrics based on formal constraints}.
\newblock \bibinfo{journal}{\emph{Information Retrieval}} \bibinfo{volume}{12},
  \bibinfo{number}{4} (\bibinfo{date}{01 Aug} \bibinfo{year}{2009}),
  \bibinfo{pages}{461--486}.
\newblock
\showISSN{1573-7659}
\urldef\tempurl%
\url{https://doi.org/10.1007/s10791-008-9066-8}
\showDOI{\tempurl}


\bibitem[\protect\citeauthoryear{Bagnall}{Bagnall}{2016}]%
        {bagnall2016authorship}
\bibfield{author}{\bibinfo{person}{Douglas Bagnall}.}
  \bibinfo{year}{2016}\natexlab{}.
\newblock \showarticletitle{Authorship clustering using multi-headed recurrent
  neural networks}.
\newblock \bibinfo{journal}{\emph{arXiv preprint arXiv:1608.04485}}
  (\bibinfo{year}{2016}).
\newblock


\bibitem[\protect\citeauthoryear{Banerjee, Dhillon, Ghosh, and Sra}{Banerjee
  et~al\mbox{.}}{2005}]%
        {banerjee2005spkmeansclustering}
\bibfield{author}{\bibinfo{person}{Arindam Banerjee},
  \bibinfo{person}{Inderjit~S Dhillon}, \bibinfo{person}{Joydeep Ghosh}, {and}
  \bibinfo{person}{Suvrit Sra}.} \bibinfo{year}{2005}\natexlab{}.
\newblock \showarticletitle{Clustering on the unit hypersphere using von
  Mises-Fisher distributions}.
\newblock \bibinfo{journal}{\emph{Journal of Machine Learning Research}}
  \bibinfo{volume}{6}, \bibinfo{number}{Sep} (\bibinfo{year}{2005}),
  \bibinfo{pages}{1345--1382}.
\newblock


\bibitem[\protect\citeauthoryear{Blei, Griffiths, Jordan, and Tenenbaum}{Blei
  et~al\mbox{.}}{2004}]%
        {Blei2004_HDP_nCRP}
\bibfield{author}{\bibinfo{person}{David~M Blei}, \bibinfo{person}{Thomas~L
  Griffiths}, \bibinfo{person}{Michael~I Jordan}, {and}
  \bibinfo{person}{Joshua~B Tenenbaum}.} \bibinfo{year}{2004}\natexlab{}.
\newblock \showarticletitle{{Hierarchical Topic Models and the Nested Chinese
  Restaurant Process}}.
\newblock \bibinfo{journal}{\emph{Advances in neural information processing
  systems}} (\bibinfo{year}{2004}), \bibinfo{pages}{17--24}.
\newblock


\bibitem[\protect\citeauthoryear{Blei, Kucukelbir, and McAuliffe}{Blei
  et~al\mbox{.}}{2017}]%
        {Blei2017VI}
\bibfield{author}{\bibinfo{person}{David~M. Blei}, \bibinfo{person}{Alp
  Kucukelbir}, {and} \bibinfo{person}{Jon~D. McAuliffe}.}
  \bibinfo{year}{2017}\natexlab{}.
\newblock \showarticletitle{Variational Inference: A Review for Statisticians}.
\newblock \bibinfo{journal}{\emph{J. Amer. Statist. Assoc.}}
  \bibinfo{volume}{112}, \bibinfo{number}{518} (\bibinfo{year}{2017}),
  \bibinfo{pages}{859--877}.
\newblock
\urldef\tempurl%
\url{https://doi.org/10.1080/01621459.2017.1285773}
\showDOI{\tempurl}
\showeprint{https://doi.org/10.1080/01621459.2017.1285773}


\bibitem[\protect\citeauthoryear{Blei, Ng, and Jordan}{Blei
  et~al\mbox{.}}{2003}]%
        {LDA}
\bibfield{author}{\bibinfo{person}{David~M Blei}, \bibinfo{person}{Andrew~Y
  Ng}, {and} \bibinfo{person}{Michael~I Jordan}.}
  \bibinfo{year}{2003}\natexlab{}.
\newblock \showarticletitle{Latent dirichlet allocation}.
\newblock \bibinfo{journal}{\emph{Journal of machine Learning research}}
  \bibinfo{volume}{3}, \bibinfo{number}{Jan} (\bibinfo{year}{2003}),
  \bibinfo{pages}{993--1022}.
\newblock


\bibitem[\protect\citeauthoryear{{Ding}, {Fung}, {Iqbal}, and {Cheung}}{{Ding}
  et~al\mbox{.}}{2019}]%
        {Ding2019Learning}
\bibfield{author}{\bibinfo{person}{S.~H.~H. {Ding}}, \bibinfo{person}{B.~C.~M.
  {Fung}}, \bibinfo{person}{F. {Iqbal}}, {and} \bibinfo{person}{W.~K.
  {Cheung}}.} \bibinfo{year}{2019}\natexlab{}.
\newblock \showarticletitle{Learning Stylometric Representations for Authorship
  Analysis}.
\newblock \bibinfo{journal}{\emph{IEEE Transactions on Cybernetics}}
  \bibinfo{volume}{49}, \bibinfo{number}{1} (\bibinfo{date}{Jan}
  \bibinfo{year}{2019}), \bibinfo{pages}{107--121}.
\newblock
\urldef\tempurl%
\url{https://doi.org/10.1109/TCYB.2017.2766189}
\showDOI{\tempurl}


\bibitem[\protect\citeauthoryear{Garc{\'\i}a-Mondeja, Castro-Castro,
  Lavielle-Castro, and Mu{\~n}oz}{Garc{\'\i}a-Mondeja et~al\mbox{.}}{2017}]%
        {garcia2017discovering}
\bibfield{author}{\bibinfo{person}{Yasmany Garc{\'\i}a-Mondeja},
  \bibinfo{person}{Daniel Castro-Castro}, \bibinfo{person}{Vania
  Lavielle-Castro}, {and} \bibinfo{person}{Rafael Mu{\~n}oz}.}
  \bibinfo{year}{2017}\natexlab{}.
\newblock \showarticletitle{Discovering Author Groups using a $\beta$-compact
  graph-based clustering.}. In \bibinfo{booktitle}{\emph{CLEF (Working Notes),
  CEUR Workshop Proceedings}}, Vol.~\bibinfo{volume}{1866}.
\newblock


\bibitem[\protect\citeauthoryear{G{\'o}mez-Adorno, Mart{\'i}n-del
  Campo-Rodr{\'i}guez, Sidorov, Alem{\'a}n, Vilari{\~{n}}o, and
  Pinto}{G{\'o}mez-Adorno et~al\mbox{.}}{2018}]%
        {gomez2018hierarchical}
\bibfield{author}{\bibinfo{person}{Helena G{\'o}mez-Adorno},
  \bibinfo{person}{Carolina Mart{\'i}n-del Campo-Rodr{\'i}guez},
  \bibinfo{person}{Grigori Sidorov}, \bibinfo{person}{Yuridiana Alem{\'a}n},
  \bibinfo{person}{Darnes Vilari{\~{n}}o}, {and} \bibinfo{person}{David
  Pinto}.} \bibinfo{year}{2018}\natexlab{}.
\newblock \showarticletitle{Hierarchical Clustering Analysis: The
  Best-Performing Approach at PAN 2017 Author Clustering Task}. In
  \bibinfo{booktitle}{\emph{Experimental IR Meets Multilinguality,
  Multimodality, and Interaction}}, \bibfield{editor}{\bibinfo{person}{Patrice
  Bellot}, \bibinfo{person}{Chiraz Trabelsi}, \bibinfo{person}{Josiane Mothe},
  \bibinfo{person}{Fionn Murtagh}, \bibinfo{person}{Jian~Yun Nie},
  \bibinfo{person}{Laure Soulier}, \bibinfo{person}{Eric SanJuan},
  \bibinfo{person}{Linda Cappellato}, {and} \bibinfo{person}{Nicola Ferro}}
  (Eds.). \bibinfo{publisher}{Springer International Publishing},
  \bibinfo{address}{Cham}, \bibinfo{pages}{216--223}.
\newblock
\showISBNx{978-3-319-98932-7}


\bibitem[\protect\citeauthoryear{Halvani, Winter, and Graner}{Halvani
  et~al\mbox{.}}{2017}]%
        {halvani2017authorship}
\bibfield{author}{\bibinfo{person}{Oren Halvani}, \bibinfo{person}{Christian
  Winter}, {and} \bibinfo{person}{Lukas Graner}.}
  \bibinfo{year}{2017}\natexlab{}.
\newblock \showarticletitle{Authorship verification based on
  compression-models}.
\newblock \bibinfo{journal}{\emph{arXiv preprint arXiv:1706.00516}}
  (\bibinfo{year}{2017}).
\newblock


\bibitem[\protect\citeauthoryear{Hamerly and Elkan}{Hamerly and Elkan}{2004}]%
        {hamerly2004learning}
\bibfield{author}{\bibinfo{person}{Greg Hamerly} {and} \bibinfo{person}{Charles
  Elkan}.} \bibinfo{year}{2004}\natexlab{}.
\newblock \showarticletitle{Learning the k in k-means}. In
  \bibinfo{booktitle}{\emph{Advances in neural information processing
  systems}}. \bibinfo{pages}{281--288}.
\newblock


\bibitem[\protect\citeauthoryear{Hong and Davison}{Hong and Davison}{2010}]%
        {Hong2010TM_Twitter}
\bibfield{author}{\bibinfo{person}{Liangjie Hong} {and}
  \bibinfo{person}{Brian~D. Davison}.} \bibinfo{year}{2010}\natexlab{}.
\newblock \showarticletitle{Empirical Study of Topic Modeling in Twitter}. In
  \bibinfo{booktitle}{\emph{Proceedings of the First Workshop on Social Media
  Analytics}} \emph{(\bibinfo{series}{SOMA '10})}. \bibinfo{publisher}{ACM},
  \bibinfo{address}{New York, NY, USA}, \bibinfo{pages}{80--88}.
\newblock
\showISBNx{978-1-4503-0217-3}
\urldef\tempurl%
\url{https://doi.org/10.1145/1964858.1964870}
\showDOI{\tempurl}


\bibitem[\protect\citeauthoryear{Kocher and Savoy}{Kocher and Savoy}{2016}]%
        {kocher2016unine}
\bibfield{author}{\bibinfo{person}{Mirco Kocher} {and} \bibinfo{person}{Jacques
  Savoy}.} \bibinfo{year}{2016}\natexlab{}.
\newblock \showarticletitle{UniNE at CLEF 2016: Author Clustering.}. In
  \bibinfo{booktitle}{\emph{CLEF (Working Notes)}}. \bibinfo{pages}{895--902}.
\newblock


\bibitem[\protect\citeauthoryear{Labovitz}{Labovitz}{1968}]%
        {Labovitz1968siglevel}
\bibfield{author}{\bibinfo{person}{Sanford Labovitz}.}
  \bibinfo{year}{1968}\natexlab{}.
\newblock \showarticletitle{Criteria for Selecting a Significance Level: A Note
  on the Sacredness of .05}.
\newblock \bibinfo{journal}{\emph{The American Sociologist}}
  \bibinfo{volume}{3}, \bibinfo{number}{3} (\bibinfo{year}{1968}),
  \bibinfo{pages}{220--222}.
\newblock
\showISSN{00031232, 19364784}
\urldef\tempurl%
\url{http://www.jstor.org/stable/27701367}
\showURL{%
\tempurl}


\bibitem[\protect\citeauthoryear{Pedregosa, Varoquaux, Gramfort, Michel,
  Thirion, Grisel, Blondel, Prettenhofer, Weiss, Dubourg, Vanderplas, Passos,
  Cournapeau, Brucher, Perrot, and Duchesnay}{Pedregosa et~al\mbox{.}}{2011}]%
        {scikit-learn}
\bibfield{author}{\bibinfo{person}{F. Pedregosa}, \bibinfo{person}{G.
  Varoquaux}, \bibinfo{person}{A. Gramfort}, \bibinfo{person}{V. Michel},
  \bibinfo{person}{B. Thirion}, \bibinfo{person}{O. Grisel},
  \bibinfo{person}{M. Blondel}, \bibinfo{person}{P. Prettenhofer},
  \bibinfo{person}{R. Weiss}, \bibinfo{person}{V. Dubourg}, \bibinfo{person}{J.
  Vanderplas}, \bibinfo{person}{A. Passos}, \bibinfo{person}{D. Cournapeau},
  \bibinfo{person}{M. Brucher}, \bibinfo{person}{M. Perrot}, {and}
  \bibinfo{person}{E. Duchesnay}.} \bibinfo{year}{2011}\natexlab{}.
\newblock \showarticletitle{Scikit-learn: Machine Learning in {P}ython}.
\newblock \bibinfo{journal}{\emph{Journal of Machine Learning Research}}
  \bibinfo{volume}{12} (\bibinfo{year}{2011}), \bibinfo{pages}{2825--2830}.
\newblock


\bibitem[\protect\citeauthoryear{Potha and Stamatatos}{Potha and
  Stamatatos}{2018}]%
        {potha2018intrinsic}
\bibfield{author}{\bibinfo{person}{Nektaria Potha} {and}
  \bibinfo{person}{Efstathios Stamatatos}.} \bibinfo{year}{2018}\natexlab{}.
\newblock \showarticletitle{Intrinsic Author Verification Using Topic
  Modeling}. In \bibinfo{booktitle}{\emph{Proceedings of the 10th Hellenic
  Conference on Artificial Intelligence}}. ACM, \bibinfo{pages}{20}.
\newblock


\bibitem[\protect\citeauthoryear{Potha and Stamatatos}{Potha and
  Stamatatos}{2019a}]%
        {potha2019dynamic}
\bibfield{author}{\bibinfo{person}{Nektaria Potha} {and}
  \bibinfo{person}{Efstathios Stamatatos}.} \bibinfo{year}{2019}\natexlab{a}.
\newblock \showarticletitle{Dynamic Ensemble Selection for Author
  Verification}. In \bibinfo{booktitle}{\emph{European Conference on
  Information Retrieval}}. Springer, \bibinfo{pages}{102--115}.
\newblock


\bibitem[\protect\citeauthoryear{Potha and Stamatatos}{Potha and
  Stamatatos}{2019b}]%
        {potha2019improving}
\bibfield{author}{\bibinfo{person}{Nektaria Potha} {and}
  \bibinfo{person}{Efstathios Stamatatos}.} \bibinfo{year}{2019}\natexlab{b}.
\newblock \showarticletitle{Improving author verification based on topic
  modeling}.
\newblock \bibinfo{journal}{\emph{Journal of the Association for Information
  Science and Technology}} (\bibinfo{year}{2019}).
\newblock


\bibitem[\protect\citeauthoryear{Raftery and Lewis}{Raftery and Lewis}{1991}]%
        {raftery1991many}
\bibfield{author}{\bibinfo{person}{Adrian~E Raftery} {and}
  \bibinfo{person}{Steven Lewis}.} \bibinfo{year}{1991}\natexlab{}.
\newblock \bibinfo{booktitle}{\emph{How many iterations in the Gibbs sampler?}}
\newblock \bibinfo{type}{{T}echnical {R}eport}.
  \bibinfo{institution}{WASHINGTON UNIV SEATTLE DEPT OF STATISTICS}.
\newblock


\bibitem[\protect\citeauthoryear{Rexha, Kr{\"o}ll, Ziak, and Kern}{Rexha
  et~al\mbox{.}}{2018}]%
        {rexha2018authorship}
\bibfield{author}{\bibinfo{person}{Andi Rexha}, \bibinfo{person}{Mark
  Kr{\"o}ll}, \bibinfo{person}{Hermann Ziak}, {and} \bibinfo{person}{Roman
  Kern}.} \bibinfo{year}{2018}\natexlab{}.
\newblock \showarticletitle{Authorship identification of documents with high
  content similarity}.
\newblock \bibinfo{journal}{\emph{Scientometrics}} \bibinfo{volume}{115},
  \bibinfo{number}{1} (\bibinfo{year}{2018}), \bibinfo{pages}{223--237}.
\newblock


\bibitem[\protect\citeauthoryear{Rosen-Zvi, Chemudugunta, Griffiths, Smyth, and
  Steyvers}{Rosen-Zvi et~al\mbox{.}}{2010}]%
        {Rosen-Zvi:2010:LAM:1658377.1658381}
\bibfield{author}{\bibinfo{person}{Michal Rosen-Zvi},
  \bibinfo{person}{Chaitanya Chemudugunta}, \bibinfo{person}{Thomas Griffiths},
  \bibinfo{person}{Padhraic Smyth}, {and} \bibinfo{person}{Mark Steyvers}.}
  \bibinfo{year}{2010}\natexlab{}.
\newblock \showarticletitle{Learning Author-topic Models from Text Corpora}.
\newblock \bibinfo{journal}{\emph{ACM Trans. Inf. Syst.}} \bibinfo{volume}{28},
  \bibinfo{number}{1}, Article \bibinfo{articleno}{4} (\bibinfo{date}{Jan.}
  \bibinfo{year}{2010}), \bibinfo{numpages}{38}~pages.
\newblock
\showISSN{1046-8188}
\urldef\tempurl%
\url{https://doi.org/10.1145/1658377.1658381}
\showDOI{\tempurl}


\bibitem[\protect\citeauthoryear{Sari and Stevenson}{Sari and
  Stevenson}{2016}]%
        {sari2016exploring}
\bibfield{author}{\bibinfo{person}{Yunita Sari} {and} \bibinfo{person}{Mark
  Stevenson}.} \bibinfo{year}{2016}\natexlab{}.
\newblock \showarticletitle{Exploring Word Embeddings and Character N-Grams for
  Author Clustering.}. In \bibinfo{booktitle}{\emph{CLEF (Working Notes)}}.
  \bibinfo{pages}{984--991}.
\newblock


\bibitem[\protect\citeauthoryear{Stamatatos}{Stamatatos}{2009}]%
        {Stamatatos2009Survey}
\bibfield{author}{\bibinfo{person}{Efstathios Stamatatos}.}
  \bibinfo{year}{2009}\natexlab{}.
\newblock \showarticletitle{A survey of modern authorship attribution methods}.
\newblock \bibinfo{journal}{\emph{Journal of the American Society for
  Information Science and Technology}} \bibinfo{volume}{60},
  \bibinfo{number}{3} (\bibinfo{year}{2009}), \bibinfo{pages}{538--556}.
\newblock
\urldef\tempurl%
\url{https://doi.org/10.1002/asi.21001}
\showDOI{\tempurl}
\showeprint{https://onlinelibrary.wiley.com/doi/pdf/10.1002/asi.21001}


\bibitem[\protect\citeauthoryear{Stamatatos, Tschnuggnall, Verhoeven,
  Daelemans, Specht, Stein, and Potthast}{Stamatatos et~al\mbox{.}}{2016}]%
        {stamatatos2016clustering}
\bibfield{author}{\bibinfo{person}{Estathios Stamatatos},
  \bibinfo{person}{Michael Tschnuggnall}, \bibinfo{person}{Ben Verhoeven},
  \bibinfo{person}{Walter Daelemans}, \bibinfo{person}{Gunther Specht},
  \bibinfo{person}{Benno Stein}, {and} \bibinfo{person}{Michael Potthast}.}
  \bibinfo{year}{2016}\natexlab{}.
\newblock \showarticletitle{Clustering by authorship within and across
  documents}. In \bibinfo{booktitle}{\emph{Working Notes Papers of the CLEF
  2016 Evaluation Labs. CEUR Workshop Proceedings/Balog, Krisztian [edit.]; et
  al.}} \bibinfo{pages}{691--715}.
\newblock


\bibitem[\protect\citeauthoryear{Teh, Jordan, Beal, and Blei}{Teh
  et~al\mbox{.}}{2006}]%
        {Yee2006HDP}
\bibfield{author}{\bibinfo{person}{Yee~Whye Teh}, \bibinfo{person}{Michael~I
  Jordan}, \bibinfo{person}{Matthew~J Beal}, {and} \bibinfo{person}{David~M
  Blei}.} \bibinfo{year}{2006}\natexlab{}.
\newblock \showarticletitle{Hierarchical Dirichlet Processes}.
\newblock \bibinfo{journal}{\emph{J. Amer. Statist. Assoc.}}
  \bibinfo{volume}{101}, \bibinfo{number}{476} (\bibinfo{year}{2006}),
  \bibinfo{pages}{1566--1581}.
\newblock
\urldef\tempurl%
\url{https://doi.org/10.1198/016214506000000302}
\showDOI{\tempurl}
\showeprint{https://doi.org/10.1198/016214506000000302}


\bibitem[\protect\citeauthoryear{Tibshirani, Walther, and Hastie}{Tibshirani
  et~al\mbox{.}}{2001}]%
        {tibshirani2001estimating}
\bibfield{author}{\bibinfo{person}{Robert Tibshirani},
  \bibinfo{person}{Guenther Walther}, {and} \bibinfo{person}{Trevor Hastie}.}
  \bibinfo{year}{2001}\natexlab{}.
\newblock \showarticletitle{Estimating the number of clusters in a data set via
  the gap statistic}.
\newblock \bibinfo{journal}{\emph{Journal of the Royal Statistical Society:
  Series B (Statistical Methodology)}} \bibinfo{volume}{63},
  \bibinfo{number}{2} (\bibinfo{year}{2001}), \bibinfo{pages}{411--423}.
\newblock


\bibitem[\protect\citeauthoryear{Tschuggnall, Stamatatos, Verhoeven, Daelemans,
  Specht, Stein, and Potthast}{Tschuggnall et~al\mbox{.}}{2017}]%
        {tschuggnall2017overview}
\bibfield{author}{\bibinfo{person}{Michael Tschuggnall},
  \bibinfo{person}{Efstathios Stamatatos}, \bibinfo{person}{Ben Verhoeven},
  \bibinfo{person}{Walter Daelemans}, \bibinfo{person}{G{\"u}nther Specht},
  \bibinfo{person}{Benno Stein}, {and} \bibinfo{person}{Martin Potthast}.}
  \bibinfo{year}{2017}\natexlab{}.
\newblock \showarticletitle{Overview of the author identification task at
  PAN-2017: style breach detection and author clustering}. In
  \bibinfo{booktitle}{\emph{Working Notes Papers of the CLEF 2017 Evaluation
  Labs/Cappellato, Linda [edit.]; et al.}} \bibinfo{pages}{1--22}.
\newblock


\bibitem[\protect\citeauthoryear{Vinh, Epps, and Bailey}{Vinh
  et~al\mbox{.}}{2010}]%
        {vinh2010information}
\bibfield{author}{\bibinfo{person}{Nguyen~Xuan Vinh}, \bibinfo{person}{Julien
  Epps}, {and} \bibinfo{person}{James Bailey}.}
  \bibinfo{year}{2010}\natexlab{}.
\newblock \showarticletitle{Information theoretic measures for clusterings
  comparison: Variants, properties, normalization and correction for chance}.
\newblock \bibinfo{journal}{\emph{Journal of Machine Learning Research}}
  \bibinfo{volume}{11}, \bibinfo{number}{Oct} (\bibinfo{year}{2010}),
  \bibinfo{pages}{2837--2854}.
\newblock


\bibitem[\protect\citeauthoryear{Wagstaff, Cardie, Rogers, Schr{\"o}dl,
  et~al\mbox{.}}{Wagstaff et~al\mbox{.}}{2001}]%
        {wagstaff2001constrained}
\bibfield{author}{\bibinfo{person}{Kiri Wagstaff}, \bibinfo{person}{Claire
  Cardie}, \bibinfo{person}{Seth Rogers}, \bibinfo{person}{Stefan Schr{\"o}dl},
  {et~al\mbox{.}}} \bibinfo{year}{2001}\natexlab{}.
\newblock \showarticletitle{Constrained k-means clustering with background
  knowledge}. In \bibinfo{booktitle}{\emph{Icml}}, Vol.~\bibinfo{volume}{1}.
  \bibinfo{pages}{577--584}.
\newblock


\bibitem[\protect\citeauthoryear{Wang and Blei}{Wang and Blei}{2012}]%
        {wang2012split}
\bibfield{author}{\bibinfo{person}{Chong Wang} {and} \bibinfo{person}{David~M.
  Blei}.} \bibinfo{year}{2012}\natexlab{}.
\newblock \bibinfo{title}{A Split-Merge MCMC Algorithm for the Hierarchical
  Dirichlet Process}.
\newblock
\newblock
\showeprint[arxiv]{stat.ML/1201.1657}


\bibitem[\protect\citeauthoryear{{Whye Teh}, Jordan, Beal, Blei, and {Whye
  TEH}}{{Whye Teh} et~al\mbox{.}}{2006}]%
        {WhyeTeh2006_HDPs}
\bibfield{author}{\bibinfo{person}{Yee {Whye Teh}}, \bibinfo{person}{Michael~I
  Jordan}, \bibinfo{person}{Matthew~J Beal}, \bibinfo{person}{David~M Blei},
  {and} \bibinfo{person}{Yee {Whye TEH}}.} \bibinfo{year}{2006}\natexlab{}.
\newblock \showarticletitle{{Hierarchical Dirichlet Processes}}.
\newblock \bibinfo{journal}{\emph{J. Amer. Statist. Assoc.}}
  \bibinfo{volume}{101} (\bibinfo{year}{2006}), \bibinfo{pages}{476--1566}.
\newblock
\urldef\tempurl%
\url{https://doi.org/10.1198/016214506000000302doi.org/10.1198/016214506000000302}
\showDOI{\tempurl}


\bibitem[\protect\citeauthoryear{Yan, Guo, Lan, and Cheng}{Yan
  et~al\mbox{.}}{2013}]%
        {YanBTM2013}
\bibfield{author}{\bibinfo{person}{Xiaohui Yan}, \bibinfo{person}{Jiafeng Guo},
  \bibinfo{person}{Yanyan Lan}, {and} \bibinfo{person}{Xueqi Cheng}.}
  \bibinfo{year}{2013}\natexlab{}.
\newblock \showarticletitle{A Biterm Topic Model for Short Texts}. In
  \bibinfo{booktitle}{\emph{Proceedings of the 22Nd International Conference on
  World Wide Web}} \emph{(\bibinfo{series}{WWW '13})}.
  \bibinfo{publisher}{ACM}, \bibinfo{address}{New York, NY, USA},
  \bibinfo{pages}{1445--1456}.
\newblock
\showISBNx{978-1-4503-2035-1}
\urldef\tempurl%
\url{https://doi.org/10.1145/2488388.2488514}
\showDOI{\tempurl}


\end{thebibliography}

\end{document}